# Failure Modes in LLM Systems: A System-Level Taxonomy for Reliable AI Applications

Vaishali Vinay
vpapneja@microsoft.com
Microsoft Security Research
Redmond, Washington, USA

*Abstract*—Large language models (LLMs) are being rapidly integrated into decision-support tools, automation workflows, and AI-enabled software systems. However, their behavior in production environments remains poorly understood, and their failure patterns differ fundamentally from those of traditional machine learning models. This paper presents a system-level taxonomy of fifteen hidden failure modes that arise in real-world LLM applications, including multi-step reasoning drift, latent inconsistency, context-boundary degradation, incorrect tool invocation, version drift, and cost-driven performance collapse. Using this taxonomy, we analyze the growing gap in evaluation and monitoring practices: existing benchmarks measure knowledge or reasoning but provide little insight into stability, reproducibility, drift, or workflow integration. We further examine the production challenges associated with deploying LLMs—including observability limitations, cost constraints, and update-induced regressions—and outline high-level design principles for building reliable, maintainable, and cost-aware LLM systems. Finally, we outline high-level design principles for building reliable, maintainable, and cost-aware LLM-based systems. By framing LLM reliability as a system-engineering problem rather than a purely model-centric one, this work provides an analytical foundation for future research on evaluation methodology, AI system robustness, and dependable LLM deployment.

*Keywords*—Large language Models, LLM systems, system-level taxonomy, failure modes, reliability, multi-step reasoning, AI reliability

## I. INTRODUCTION

Large language models (LLMs) based applications and AI agents are playing a significant role in modern information systems, decision-support pipelines, and enterprise automation architectures. This capability for understanding heterogeneous data sources, organizing multi-step responsibilities, and interfacing with external tools has sped up their deployment in areas such as healthcare, finance, education, and cybersecurity [1], [2], [3], [4]. Despite all the progress, the overall reliability issues are known to affect the systems, but these are not limited to isolated errors in model output, but they arise from system-level failure modes which are often hidden, interacting, and challenging to detect with the existing testing. These failure modes could remain unnoticed during successful demonstration but could emerge under realistic operating conditions, which would reveal behavioral weaknesses that are not visible in the early stages. The challenge of ensuring that these systems can and will deliver credible responses, especially post-deployment, persists despite these advances. Early testing contexts exhibit impressive performance in practice; however, these evaluations in practice usually do not capture the operational conditions under which agents need to work with respect to reliability and consistency. A burgeoning literature shows that LLM output remains highly variable with repeated run times that use the same type of cues. Output divergence in multi-step reasoning tasks has been reported to be g$^1$reater than 20–30%, and with inconsistent intermediate steps, the final answer can also be unstable [5]. This level of variability is even greater for long-horizon tasks, where minor deviations in intermediate reasoning accumulate into significant behavioral drift. Moreover, experiments on agent-based systems also show that task-sequencing errors, silent failures, and improper tool invocation happen at non-trivial rates within actual interactions, weakening agent reliability despite strong performance in controlled experiments [6]. Figure 1 provides an overview of the failure rate of Multi-agent LLMs, and as we can see, in some models, the failure rate is more than in others [6].

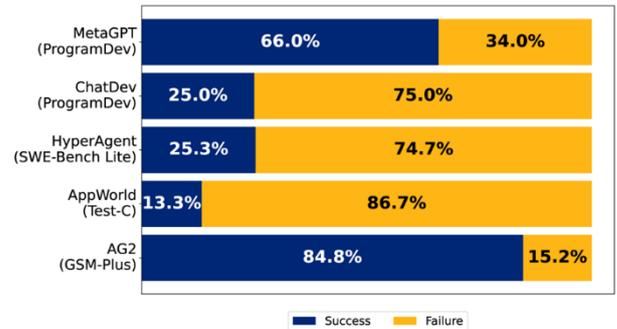

*Figure 1. Failure rate of five popular Multi-agent LLM*

Cost constraints are known to further worsen the risk, as many real-world GenAI deployments would operate within strict inference-cost budgets, which would encourage smaller models, shorter context windows, or even reduced sampling. Studies focused on inference-cost and compute-budget constraints in LLM deployments suggest that the trade-off between cost and reasoning accuracy, but explicit links between sot-driven reductions and tool-use failures are known to be an area that is underexplored [7]. Going from proof-of-concept to production deployment comes with new reliability constraints. Examples of this are changing input distributions, vague or noisy user instructions, unexpected tool latencies, and ever-changing operational environments. Studies of deployment-induced changes in behavior show that adjustments to the underlying versions of the model or inference-time parameters can introduce regression in previously stable behavior, causing instability in the predictability of the system and increasing the complexity of long-term maintenance [8]. It is critical to note that reliable evaluation is a necessity even at the proof-of-concept stage,

---

as early testing allows for ensuring accuracy or plausibility under controlled prompts that would often mask volatility, reasoning drift, and incomplete tool use coverage [9]. All of these would result in the evaluation appearing to be convincing during the prototyping, which often is not ideal in predicting long-term reliability. In multi-agent workflows, where one erroneous output runs through dependent components and produces cascading failures that are difficult to detect or analyze [10], these issues are magnified. Traditional evaluation metrics like accuracy, perplexity, or static benchmark performance fall short of representing these system-level reliability issues. These metrics are essentially about linguistic or cognitive capabilities rather than the stability, reproducibility, or long-term behavioral integrity that are essential for reliable deployment.

These evaluation limitations would motivate the need for a system-level perspective on reliability, and current taxonomies are focused on hallucinations, bias, or abstract safety risks, but they are not capable of capturing the failure mechanism that would emerge from interaction, memory, versioning, tool orchestration, or cost-induced degradation [11]. The structure taxonomy of system-level failure modes is thus critical in understanding, predicting, as well as mitigating the reliability threats across the full lifecycle of GenAI systems [12]. The current gap between pre-deployment validation and post-deployment reliability is of increasing concern for organizations, which raises the following central research question:

***How can we trust AI agents' responses not only at the first proof-of-concept, but during their lifecycle in production settings?***

To this end, this paper develops a system-level taxonomy of hidden failure modes, highlights shortcomings of existing evaluation methodologies, examines challenges of reliability inherent to the production deployment, and outlines design principles tailored to ensure stability, robustness, and trust in AI agent responses over sustained use in the field

## II. BACKGROUND

Rapid progress in language modeling has paved the way for significant advancements in natural language understanding, generation, and contextual reasoning. Many modern LLMs include summarization, code generation, multi-document synthesis, and structured task planning capabilities [13], [14]. As a result, the perception has prevailed that the principal hurdles of introducing language-based agents are associated with the enhanced quality of the models' raw cognitive performance. Yet operational experiments show that capabilities are not a guaranteed means of reliability, particularly for agents that should be capable of functioning autonomously in dynamic environments.

The distinguishing features of LLMs introduce reliability concerns that vary radically from those faced by traditional machine learning algorithms [15], [16]. The generation of this model introduces significant variability in outputs, in the same prompt or execution environment. In controlled experiments in the literature, repeated inference on deterministic prompts can yield markedly different outputs across runs, and some of these may exhibit no deterministic test outputs at all, which would complicate the evaluation and debugging process [17]. Second, prompt sensitivity leads to marked behavioral changes with minor differences in terms of input phrasing, format, or arrangement (i.e., input order) but with the same semantic intent [18]. Third, context window constraints bring degradation on extended input sequences, where the information in proximity to boundary regions is more likely to be omitted, misinterpreted, or semantically distorted [19]. Furthermore, tool-use dependency adds a different layer of system-level risk as agents must take in the observations, choose the tools, and incorporate the solutions into the remainder of their reasoning; failure is no longer solely a result of language generation, rather it is the result of interaction [20].

Such properties differ significantly from classical machine learning (ML) evaluation approaches that make assumptions of deterministic inference processes and a fixed mapping from input to output. Performance benchmarking is a strong predictor of real-world applicability in classical models, and evaluates knowledge and reasoning that occurs in idealized conditions for LLM-based agents, but does not account for long-horizon consistency, reproducibility, or integration reliability in entire software systems. Benchmark-aligned improvements often fail to translate into downstream operational stability, with deployments still exhibiting unforeseen mistakes despite pre-deployment evaluations [21], [22]. Reliability-related research has revolved chiefly around hallucinations, bias, and safety safeguards, and although these are important issues, they do not fully describe the systemic failure patterns seen in actual deployments. The failure behaviors specific to production environments, such as reasoning drift, degraded behavior based on noisy inputs, regression due to model updates, and cascading errors in multi-agent systems, are insufficiently captured by classical taxonomies.

Consequently, considerable operational problems arise later than during experimentation and after integration. This gap has helped to foster the idea that system-level safety and reliability of LLM problems should be considered system-engineering problems as opposed to just being model-centric issues. Trustworthiness should involve not only linguistic competence, but also stability under perturbation, consistency across time, and predictable interaction with the surrounding software. Understanding failure modes and evaluation gaps at the system level, therefore, is an essential step toward allowing dependable deployment of AI agents.

## III. SYSTEM-LEVEL TAXONOMY OF HIDDEN FAILURE MODES IN LLM-BASED APPLICATIONS

Here in this section, we explore the fifteen system-level failure modes observed in LLM-based applications that are categorized into three dimensions: Reasoning failures, input and context failures, and system and operational failures [23], [24]. The 15 system-level failure modes are clustered into these three dimensions, as each dimension represents a separate location in the failure spectrum of an LLM application pipeline. Reasoning failures represent errors that occur internally to the model, even when the prompt is correct, e.g., hallucinations, logical contradictions, planning collapse, and problems with calibration. These arise due to limitations of the model's internal representations and probabilistic reasoning. Prior to generation, input and context failures are apparent, arising not from model behavior but rather the brittleness of the prompt and context interface. Inducing ambiguity, prompt injection, context loss, distribution shift, and conflicting instructions all lead to performance instability independent of model quality. Lastly, system and operational faults arise post-generation in the compatibility-rich application space with tool-invocation anomalies, composition failures, business-rule misfits, multi-agent communication problems, and precision losses by compromises between cost or latency. This type of failure mode classification facilitates easier diagnosis and mitigation.

Unlike the model-centric surveys, the framing is based on how these failures emerge when LLMs are embedded within the multi-step pipelines, tools, or multi-agent workflows that are often where the reliability degradation becomes more pronounced. Figure 2 provides an overview of the classification and the 15 failure modes.

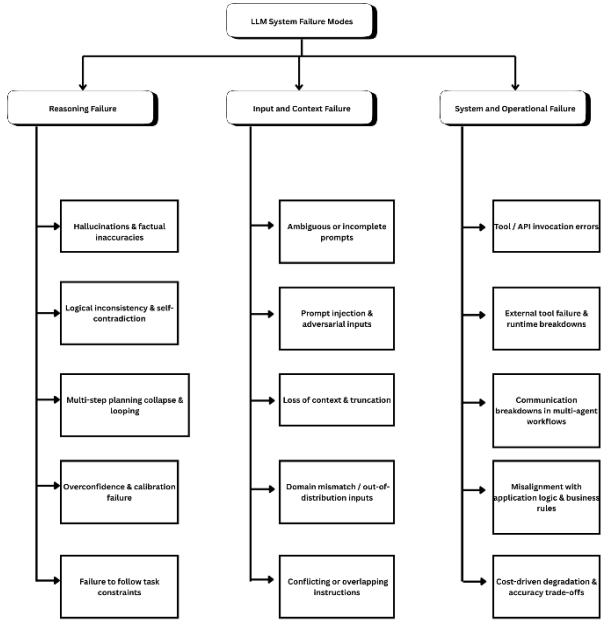

Figure 2. LLM System Failure taxonomy

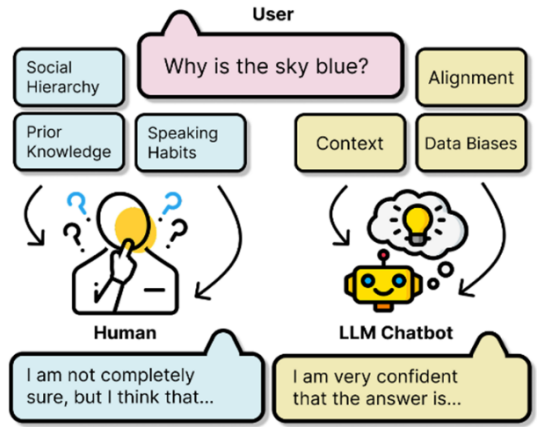

Figure 3. Human vs AI

## A. Reasoning Failures

### 1) Hallucinations & factual inaccuracies

Hallucinations and factual inaccuracies occur when LLMs produce non-factual but fluent utterances, as they maximize linguistic likelihood rather than truth; this is known as plausible yet non-factual output [6]. Particularly in system deployments, hallucinations become especially dangerous because they silently propagate across modules, downstream tools, and agents operate on fabricated information without any error signal. This is different from visible hallucinations in chat, because nothing explicitly flags the fabricated content, resulting in error amplification and unpredictable end-to-end behavior [25].

### 2) Logical inconsistency & self-contradiction

Logical inconsistency and self-contradiction represent a separate failure class [26], [27]. Many LLMs repeat earlier steps or produce content that contradicts previous turns, and this arises from the lack of global memory consistency, as each output is generated independently of past assertions [6]. In multi-agent coordination, when one agent later contradicts something it previously committed to, collaborative planning collapses, and unlike hallucination, the core flaw here is not factual inaccuracy but an unstable internal world-state, which is difficult to detect without cross-turn consistency checks [28]. Figure 3 shows the difference in thinking and how logical inconsistencies can be avoided [29].

### 3) Multi-step planning collapse & looping

Multi-step planning collapse and looping occur when chain-of-thought or tool-use workflows exceed the stable reasoning depth of the model [30], [31]. These models can stall, skip steps, or repeat work indefinitely, which is referred to as the "step repetition" pattern, and since there is no built-in failure signal, these systems tend to fail indirectly through timeouts or unexplained deadlocks [6]. The problem is different from inconsistency because the agent doesn't contradict itself; it simply never converges toward completion, making debugging difficult.

### 4) Overconfidence & calibration failure

LLMs fail with overconfidence and calibration because they rarely express uncertainty and often present incorrect assertions with authority [32], [33]. Downstream components misinterpret linguistic confidence as epistemic certainty, and as a result, validators and humans skip verification, assuming the model is correct [29]. In this mode, the text "sounds right," which suppresses error-detection triggers and causes incorrect assertions to propagate as factual elements within workflows.

### 5) Failure to follow task constraints

The task constraints are not being adhered to because of "disobey task specification" behavior [6], which is often triggered when high-level instructions conflict with context or system feedback. The model is not intentionally disobeying instructions but drifts toward an inferred objective that diverges from the user's intent, and the system designers cannot detect failure by examining isolated responses, because partial compliance, like correct reasoning but wrong formatting, does not surface until execution time [34].

## B. Input and Context Failures

### 1) Ambiguous or incomplete prompts

Ambiguous and partial prompts cause a chain reaction of failure, as the model assumes a single interpretation without seeking further explanations, echoing the "fail to ask for clarification" pattern [6], [35], [36]. While the output seems logical, it encodes the incorrect comprehension, and this becomes a seed for later reasoning or the use of tools, and this mistake is frequently misdiagnosed as incorrect reasoning when, really, confusion is inherent in the input layer.

### 2) Prompt injection & adversarial inputs

Adversarial and malicious prompt injection occurs when a user injects hidden instructions into untrusted text, causing models to override normal behavior [37], [38], [39]. The weakness of many LLM pipelines is that they merely concatenate user text into prompts to allow attackers to take over agent behavior, extract user-sensitive data, or induce undesired execution. This failure, unlike ambiguous prompts, is an intentional manipulation of safety layers to convert prompt ingestion into an attack surface.

*3) Loss of context & truncation*

Loss of context and truncation occur when earlier conversation history in the context window is pushed out of the window, leading to a "loss of conversation history" [6], [40]. The system behaves as though instructions never existed, due to the model silently dropping memory, and the users notice a stark personality or goal change, which gives the appearance of random behavior. There is no failure signal produced because, from the model's view, the ignored text just does not exist.

*4) Domain mismatch / out-of-distribution inputs*

Domain mismatch and distribution shift occur when inputs come from unfamiliar or highly specialized fields, and research notes that broad pre-training "presents distinct challenges" when encountering novel domains [6], [41], [42]. These failures differ from hallucination as responses remain internally coherent but aligned to the wrong domain, producing shallow or irrelevant content, and in deployment, this causes accuracy collapse when real-world inputs differ from benchmark prompts.

*5) Conflicting or overlapping instructions*

Conflicting or overlapping instructions trigger "task derailment" [6], as the model oscillates between incompatible objectives (e.g., brevity vs. detail) [43], [44]. The system output becomes unstable, not due to reasoning faults but due to contradictory supervisory signals, and engineering teams often misattribute this to model unreliability, when the true problem lies in unresolved conflict in the control layer.

*C. System and Operational Failures*

*1) Tool / API invocation errors*

Tools/API invocation errors occur due to models generating syntactically convincing function names but not true existent ones, or invalid arguments, and you can catch this type of error when you are saying "select a tool that does not exist" [24]. Natural language reasoning does not always ensure API rules, and errors can be seen at runtime, and this failure is costly since it causes developers to diagnose the outside system rather than the model output, generating recursive timeout loops. Figure 4 provides an overview of the tool failures in Gorilla LLM, showcasing them in different scenarios [24].

*2) External tool failure & runtime breakdowns*

External tool failure and runtime problems happen even when the call itself is on track; APIs can fail, schema may change, data types may shift, rates may be pushed beyond the limit, and so on [24], [45], [46]. Pipelines tie several tools together; thus, a single runtime error gets passed on downstream, and such failures are misattributed to model logic when the issue that is being addressed is tool instability. The result is propagated outwards due to a series of faults, particularly in the presence of concurrency.

*3) Communication breakdowns in multi-agent workflows*

In multi-agent workflows, communication breakdowns translate into "conversation reset" [6], in which shared memory disappears or is overwritten in the middle of a task. Agents persist in execution but on different state representations, and unlike hallucination or inconsistency, this inability is infrastructural rather than cognitive failure, and thus is detectable only with secondary symptoms, such as a decrease in task completion rate or repeated attempts.

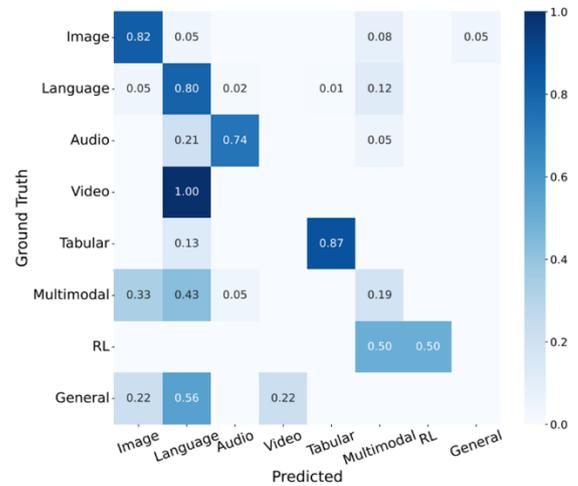

*Figure 4. Incorrect tool failures in Gorilla*

*4) Misalignment with application logic & business rules*

When model outputs follow instructions but violate domain constraints or business rules, the result is misalignment with application logic, and it can be seen from industry analyses that many failures in deployment originate not from hallucination but rather the semantic mismatch between natural-language output and software requirements [47]. Unlike tool invocation failure, outputs are syntactically valid but semantically incompatible, thus leading to silent failure at the application layer.

*5) Cost-driven degradation & accuracy trade-offs*

With systems optimizing for lower compute, cost-driven degradation and accuracy trade-offs become clear in the process. Approaches like FrugalGPT/ThriftLLM indicate that there is a way to maintain accuracy with some intelligent routing, but uncontrolled token truncation, fallback to weaker models, or aggressive caching uniformly degrade correctness and do so without triggering alerts [48], [49]. In practice, users will identify quality degradation long before monitoring systems recognize it, so accuracy degradation is a consequence of engineering choices rather than an inherent property of "agentic AI."

*D. Perspective Summary*

In all three groups, the key lesson is that these breakdowns are indeed not model failures but system failures. LLM-enabled architectures are unworkable if the fluent text hides the flawed or imprecise computation, and the detection fails too when errors appear polished in appearance, when pipelines time out instead of crashing, when the system looks "almost correct," or when the source of error is several modules upstream. As a result, scale-friendly reliability is fragile at best, evaluation gaps arise with benchmarks that fail to account for real behavior, observability requirements are high, and trust is misplaced in confident-sounding responses. For these reasons, mitigation needs to be system-design-level, not only prompt‑engineering, nor model-tuning-level, to work. Consistent state representations, formal schemas, controlled tool interfaces, disambiguation protocols, safe prompt boundaries, memory hygiene, and explicit accuracy cost restrictions are requirements to secure robust behavior during practical application.

## IV. THE EVALUATION GAP IN LLM SYSTEMS

Most of the assessments of large language models (LLMs) are still anchored to static benchmarks for their knowledge recall or task performance rather than model stability and operational reliability. For instance, in a lot of benchmarks, accuracy is evaluated on fixed test-sets, but the performance of output behavior under repeated runs, prompt perturbations, or time is not assessed. As one study explains, benchmark metrics such as BLEU, ROUGE, or accuracy "do not necessarily reflect human judgment" or behavioural reliability within open-ended systems [50].

The literature identifies another glaring limitation, the absence of ground truth in open-ended tasks (summaries, dialogues, reasoning). In cases that use human-annotated or LLM-judge ratings, they are biased and unstable [51]. A peer-reviewed meta-evaluation reported that an average of 48.4% of LLM-as-judge pipelines reversed verdicts in mirrored response order, with judges agreeing at high rates, suggesting severe instability of the evaluation mechanism itself [51].

Another issue is non-determinism. Recent examples in NAACL demonstrate that evaluations that disregard the stochastic nature of the model may lead to inaccurate conclusions, as a single output per prompt is not sufficient to capture the variability within the model behavior [52]. Standard software testing assumes a deterministic output, and so, many evaluation frameworks do not factor in run-to-run variability, which is an important consideration when LLMs are introduced in pipelines.

Finally, there is a lack of standard metrics or processes for drift, consistency, or cross-version stability. Although survey work on LLM-agent evaluation has begun to emphasize "reliability" and "long-horizon interaction" as essential dimensions, current benchmarks do not yet have operational workarounds to capture how the behavior of a model evolves or responds to internal changes [53]. Together, these challenges argue that current evaluation frameworks often overlook how the LLM systems behave when they are deployed within real workflows. Here, the focus is on what is easy instead of what is required for reliability.

A consequence is silent regression, which means if a model upgrades or changes its settings, it may degrade behavior in untested scenarios without any visible signal in benchmark metrics. These regressions may be lost to evaluation because there are no repeatability checks and drift tracking until they occur in production. Another outcome is output variability: minor changes to the phrasing of a prompt, the ordering of the data that you use, or the configuration of the system will create drastically different outputs. One study, the aforementioned on non-determinism, provides the significant variance in outcomes when a single run is used as a benchmark [52], and meta-evaluators concluded that nearly 50% of pairs of comparisons flipped upon reversal in response order [54]. This volatility lowers user trust and makes it difficult to upgrade or use A/B testing strategies.

Finally, the next implication is hard-to-debug errors. If evaluation misses the nuance of how models change over time and how behavior varies, failure diagnosis becomes anecdotal. Engineers are not able to be confident that any sort of failure is due to a prompt change, updating the model, a change in the selected version, a variation in the sampling, or a drift of the data. Root-cause analysis is significantly hampered without repeatable evaluation logs and traceability. The end result is unreliable workflows at scale because of these gaps, or organizations may select a model simply through benchmark scores, only to deploy it in a suite of tools or a multi-agent pipeline where its behavior is degrading, or shifts unpredictably. Evaluation approaches have not been shown to consistently incorporate stability considerations, drift, or real-world variability; this gap may introduce brittleness during deployment. In brief, this evaluation gap may pose operational risks when deployed.

These concerns illustrate that the performance at the surface level of task accuracy does not reveal the risks when LLM is integrated into actual workflows. Such a taxonomy, which differentiates reasoning, input/context, and system-level failure modes, is necessary as it mirrors the way in which problems appear when deploying, not only in benchmark situations. Such a structure enables methods of evaluation that evaluate not just correctness, but also stability, repeatability, drift, and alignment with downstream systems.

## V. DEPLOYMENT REALITIES: THE PRODUCTION GAP

The realities of deployment demonstrate enormous production gaps between lab-scale LLM systems and real-world operation. Version drift and model updates are among the leading factors that cause this issue; for instance, models that look stable in benchmark tests may show behavior changes when an update is performed or a provider retunes a model version, resulting in breaking changes such as changing format, reasoning style, or tool-call ordering [55]. The drift introduces the risk of regression; a workflow that has previously given steady and reasonable estimates can suddenly degrade as you simply do not alter any bit of code.

Academic work on reproducibility provides a similar documentation: e.g., Herrera-Poyatos et al. (2025), model variance and uncertainty are both still significant when holding prompts and inputs constant, so the uncertainty and randomness in updates further heighten unpredictability [56]. Due to the fact that typical software engineering practices assume deterministic, stable behavior, LLMs pose a challenge to such assumptions and leave systems open to sudden collapses.

A second issue is observability and monitoring deficiencies, and this is due to the fact that, unlike legacy software, LLMs do not expose distinct internal state, decision logs, and confidence metrics that can give the appearance of syntactical correctness but can actually be semantically incorrect or out of sync. Lots of tools already exist for monitoring infrastructure (latency, memory, errors), but none include correctness, hallucinations, drift, or tool loop inefficiencies. Thus, in its reproducibility study, the researcher notes that stochastic outputs and changing prompts complicate auditing and tracing, which means in many instances we simply don't have the telemetry necessary to question "why did this answer change?" [57].

Drift signals, prompt template modifications, context-window truncation statistics, and multi-agent call counts are infrequent in monitoring systems, and the end result is that errors can remain undetected until user implications are discovered or tools downstream break down. In LLM scenarios, drift is the evolution of model behavior over time in the absence of any intentional code changes. Version drift can be defined as the state in which a provider updates or retunes a model and causes previously stable workflows to alter their format, reasoning style, or tool-call patterns. The problem of data drift occurs when the distribution of real inputs deviates from what the model was trained or validated on, leading to a decrease in accuracy despite the fact that the actual model itself has not changed. Behavior drift occurs when the same prompt has a different output over time as a result of stochastic sampling or undocumented internal changes. Drift nullifies predictability and adds complexity to monitoring and reproducibility

Cost and latency constraints make this more complicated. Many production LLM pipelines evolve far beyond simple prompt–response patterns as chains of tool calls, retrieval-augmented generation, agent orchestration, and larger context windows increase both compute cost and latency. Token explosion (more extended conversations, multi-step reasoning) leads to spending growth and may trigger budget pressures.

To combat cost, teams frequently trim context, reduce sampling size, or simplify tool loops, which can impact performance or robustness without being highlighted by standard monitoring. Although peer-reviewed literature on LLM cost cascades is relatively new, studies of the stability [56] and reproducibility [57] of the LLM indirectly demonstrate that altering operational parameters (e.g., the sampling rate, context length) does significantly impact output quality. The engineering consequence is that these cost-optimization choices can trade accuracy or reliability for cost savings, introducing the possibility of invisible performance impacts.

Reproducibility and audit gaps threaten compliance and safety and are the final source of trouble for legality, and when an LLM-based workflow generates different outputs for a given input because of stochastic sampling, changing prompts, or model revision, past decisions cease to be reproducible. Without versioned prompts, retrieval logs, context snapshots, and tool-call traces, you can never reliably repeat why a particular result occurred weeks or months later.

The importance of this issue is emphasized by the "Analyst-Inspector" framework of Zeng et al. (2025), which emphasizes the fact that LLM-generated data science workflows are often non-reproducible and not replicable, which are damaging to transparency and trust [57]. In more regulated domains like finance, healthcare, or legal, this diminishes auditability as organizations might not be able to demonstrate how a decision was reached or whether a safety filter was consistently applied.

Collectively, these production-engineering failures demonstrate that successful lab performance does not meet the need for deployment at scale of LLM systems. Reliability does not just derive from the models' accuracy and metric scores; there are system design factors: version control, semantic monitoring, cost-accuracy governance, reproducible traces, and detecting behavioral drift, and without incorporating these capabilities into the deployment pipeline, organizations may end up with fluent but brittle systems in which even a moment of correct output can hide profound instability, cost blowouts, or compliance failures.

## VI. Design Principles for Reliable LLM-Based Systems

Systems using LLMs require architectural controls around model behavior rather than only focusing on model accuracy. Studies on prompt engineering and evaluation have shown a measurable effect on output reliability due to input consistency. One such study demonstrates that standardized prompt formats and modular prompt components contribute to response stability across multiple types of tasks, exceeding ad hoc free-form instructions in all benchmark settings [58]. Similarly, another study showed that canonical prompt patterns lead to less ambiguity and harmful variance by limiting the search space within which alternative responses can be identified [59], and this evidence reinforces the central design principle that LLM inputs need to be canonicalized, which means they would be reformatted, reordered, and de-noised before inference, and workflows need to be based on versioned prompt templates, not on dynamically assembled instructions. Figure 5 showcases how prompt engineering would work and the improvements that can be made [61].

A second pillar has to do with validation mechanisms. Although research on dedicated "verifier layers" is still emerging and not yet peer-reviewed, the evaluation literature points to the fact that undetected hallucinations and inconsistency hinder downstream system reliability. One study explains that hallucination remains difficult to detect without explicit verification and that intermediate validation can limit the spread of incorrect reasoning in multi-stage LLM pipelines [60]. Given these data, it is clear that a solid system won't consider generation the final answer, but that additional intermediate checks, like schema validation for structured outputs or reruns for consistency, should govern whether the output will go through.

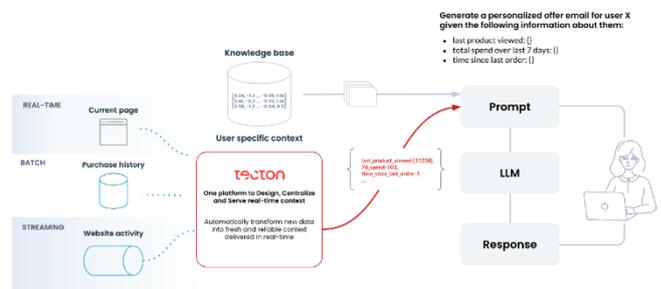

*Figure 5. Prompt Engineering*

Monitoring is also crucial for operational reliability. The vast majority of production telemetry systems monitor latency and error rates; however, they do not track behavioral drift. One empirical study indicates that data drift in ML systems can be empirically detected through longitudinal monitoring of distribution shifts, which should serve to encourage drift-aware monitoring mechanisms in AI workflows to prevent silent quality degradation [62]. This implies that accurate LLM monitoring requires output-variance tracking, formatting-change detection, and longitudinal sampling of behavioral indicators, as opposed to only assessing infrastructure health.

## VII. Future Work

Constant reliance on LLMs suggests the need for the researchers to shift the emphasis from model-centric utility to system-centric reliability. Second, a more targeted direction is to move towards more standardized measures of LLM reliability, as evaluation approaches commonly stress accuracy on fixed tasks, with limited information about their stability, repeatability, or drift. Common reliability metrics such as model consistency across prompts, robustness during paraphrasing, and stability across model versions could promote meaningful comparisons across models and deployment contexts.

Another critical area is creating benchmarks that enable multi-step reasoning drift, and today's leaderboards assess mostly isolated question-answering or small reasoning tasks, but real deployments require long-horizon planning, tool chains, and agent interactions. These, however, would better serve the production-grade behavior we would want to see in our tasks, the way LLMs maintain goals, and keep up with instructions across many steps, or avoid looping or derailment. There is also still a need to study drift detection on LLM outputs, for safety and for enterprise reliability. Systems must be able to pick up when the models subtly change formatting, style, refusal patterns, confidence, or safety posture, even when benchmarking continues to be very high. This space intersects seamlessly with anomaly detection and monitoring. Another promising direction is tool-use reliability.

Modern LLM systems often depend heavily on APIs, retrieval, and external services. Systematic metrics are required to examine if the model calls tools properly, with the appropriate arguments, in the correct order, and without hallucinating capabilities that are not there. In parallel, observability frameworks relevant to LLMs should also be investigated. Conventional infrastructure telemetry does not disclose alignment errors, hallucinations, or logical mismatches, and new signals and dashboards should thus reveal semantic reliability, rather than just uptime. Cost-performance modeling has so far been neglected. With the growth of LLM systems, the reliability of the proposed method must be maintained without unmanaged increases in latency and expenditure.

In addition, future work will need to investigate how accuracy, safety, and consistency are influenced by token budgets, model selection, and multi-step inference. Collectively, these directions indicate towards the emergence of LLM engineering as a reliability-driven discipline and not a demo-driven one.

## VIII. Conclusion

This work proposed a system-level framing and taxonomy of failure modes for LLM applications, demonstrating that failures do not typically result from a single incorrect generation. Instead, they arise from a confluence of reasoning failures, input and context volatility, and operational deficits. The taxonomy included fifteen unique failure patterns, such as hallucinations and multi-step planning collapse; tool-use faults; and version drift in a multi-stage, retrieval pipeline, or agent-based orchestration (versus an isolated prompt–response) setting.

These results redefine LLM reliability rather than a model-specific problem in systems terms. A model that performs well on fixed benchmarks may have erratic behavior as prompts are changed, components break, and costs begin to weigh on architecture decisions. This gap between laboratory accuracy and production reliability remains because we do not have performance metrics for the stability, drift, reproducibility, and cost. Progress in the field will need tools such as input canonicalization, verification layers, semantic observability, controlled versioning, and cost governance to enable trustworthy scaling.